\documentclass[runningheads]{llncs}
\usepackage{graphicx}
\usepackage{tikz}
\usepackage{comment}
\usepackage{amsmath,amssymb} % define this before the line numbering.
\usepackage{color}

\usepackage{caption}
\usepackage{subcaption}

\usepackage[accsupp]{axessibility}  % Improves PDF readability for those with disabilities.

\usepackage{booktabs}

\usepackage{float}

\begin{document}
\raggedbottom
% \linenumbers
\pagestyle{headings}
\mainmatter
\def\ECCVSubNumber{11}  % Insert your submission number here

\title{Mixed-domain Training Improves Multi-Mission Terrain Segmentation} % Replace with your title

\begin{comment}
\titlerunning{ECCV-22 submission ID \ECCVSubNumber} 
\authorrunning{ECCV-22 submission  ID \ECCVSubNumber} 
\author{Anonymous ECCV submission}
\institute{Paper ID \ECCVSubNumber}
\end{comment}
%******************

%\begin{comment}
\titlerunning{Multi-Mission Segmentation}
%\begin{comment}
\author{Grace Vincent\inst{1,2}\and
Alice Yepremyan\inst{2}\and
Jingdao Chen\inst{3}\and
Edwin Goh\inst{2}
}
%\end{comment}
%
\authorrunning{G. Vincent et al.}
%\begin{comment}
\institute{Electrical and Computer Engineering, North Carolina State University \and
Jet Propulsion Laboratory, California Institute of Technology\and
Computer Science and Engineering, Mississippi State University \\
\email{gmvincen@ncsu.edu}
\email{alice.r.yepremyan@jpl.nasa.gov}
\email{chenjingdao@cse.msstate.edu}
\email{edwin.y.goh@jpl.nasa.gov}}
%\end{comment}
%******************
\maketitle

\begin{abstract}
Planetary rover missions must utilize machine learning-based perception to continue extra-terrestrial exploration with little to no human presence. Martian terrain segmentation has been critical for rover navigation and hazard avoidance to perform further exploratory tasks, e.g. soil sample collection and searching for organic compounds. Current Martian terrain segmentation models require a large amount of labeled data to achieve acceptable performance, and also require retraining for deployment across different domains, i.e. different rover missions, or different tasks, i.e. geological identification and navigation. This research proposes a semi-supervised learning approach that leverages unsupervised contrastive pretraining of a backbone for a multi-mission semantic segmentation for Martian surfaces.
This model will expand upon the current Martian segmentation capabilities by being able to deploy across different Martian rover missions for terrain navigation, by utilizing a mixed-domain training set that ensures feature diversity.
Evaluation results of using average pixel accuracy show that a semi-supervised mixed-domain approach improves accuracy compared to single domain training and supervised training by reaching an accuracy of 97\% for the Mars Science Laboratory's Curiosity Rover and 79.6\% for the Mars 2020 Perseverance Rover.
Further, providing different weighting methods to loss functions improved the models correct predictions for minority or rare classes by over 30\% using the recall metric compared to standard cross-entropy loss.
These results can inform future multi-mission and multi-task semantic segmentation for rover missions in a data-efficient manner.

\end{abstract}

\section{Introduction}

Planetary rover missions such as Mars 2020 (M2020) \cite{williford2018nasa} and the Mars Science Laboratory (MSL) \cite{grotzinger2012mars} have been critical for the exploration of the Martian surfaces without a human presence. Terrain segmentation has been necessary for a rovers ability to navigate and maneuver autonomously, and it involves identifying features across the surface's terrain e.g., soil, sand, rocks, etc.
Automated pipelines that utilize deep learning (DL) can enable the efficient (pixel-wise) classification of large volumes of extra-terrestrial images \cite{wagstaff2021mars} for use in downstream tasks and analyses.
DL is notoriously sample inefficient \cite{li2021efficient}, often requiring on the order of 10,000 examples to achieve good performance or retraining when moving across different domains (e.g., different missions, planets, seasons lighting conditions, etc.) \cite{swan2021ai4mars}.
%conducting these automated tasks necessitates an extensive amount of data to be captured by these missions.
%To utilize this extensive data for traditional supervised DL, significant annotation and analysis effort is necessary of the collected images. 
As such, significant annotation campaigns are required to generate a large number of labeled images for model training. Furthermore, specialized annotations in the planetary science domain require the knowledge of experts such as geologists and rover drivers, which incurs additional cost. Inconsistencies with the taxonomy and lighting further complicate this issue. 
%However, DL is notoriously sample inefficient, often requiring on the order of 10,000 examples to achieve good performance as well as additional retraining when moving between different domains (e.g.,  different missions, planets, seasons, lighting conditions, etc.). 

To address this circular problem of requiring annotated examples to develop an automated annotation pipeline, this research aims to leverage semi-supervised learning for semantic segmentation (i.e., pixel-wise classification), with limited annotated examples across multiple missions. This framework will utilize a large number of unlabeled images through contrastive pretraining \cite{chen2020simple,chen2020big} for a data-efficient way to extract general features and has been shown to outperform supervised learning \cite{goh2022mars}. This backbone model is then finetuned using a mixed-domain dataset that captures the diverse feature information across the different planetary missions. The ability to achieve state-of-the-art semantic segmentation performance with minimal labels will be integral for conducting downstream science (geological identification) and engineering (navigation) tasks across the M2020 and MSL missions, as well as future Lunar and Ocean Worlds missions.

In summary, the contributions of this work are 1) analyzing a mixed-domain training set for multi-mission segmentation that isolates the change in domain and the change in number of samples to understand the effects of different training set compositions, 2) introducing a contrastive self-supervised pretraining network to improve performance on limited-label multi-mission terrain segmentation, and 3) addressing the class imbalance through reweighting loss functions.

% Contribution 1. multi-mission - combined datasets isn't novel, but we're adding to the understanding regarding mixed-domain training; isolated change in domain from change in num. labels
% Contribution 2. introduce self-supervised learning SimCLR - improves performance on label-limited M2020 dataset by 5% over published baseline; exceeds published baseline performance with only 20% of the labels
% Contribution 3. addressing class imbalance with different loss functions

% TODO: Add an overview of the paper (see pg. 2 of https://www.sciencedirect.com/science/article/pii/S001021801830511X)

The rest of the paper is organized as follows. Section \ref{sec:related} describes current computer vision applications in planetary science as well as the transfer learning methods, while Section \ref{sec:methods} details the data collection of the AI4Mars datasets and summarizes the technical approach. Section \ref{sec:result} presents in detail the results of the contributions listed above.

\section{Related Work}
\label{sec:related}

\subsection{Planetary Computer Vision}
Planetary computer vision applications have increased in the recent decades to increase capability and automation of planetary exploration and data analysis. The Planetary Data System (PDS) allows for the archival of image data collected from different NASA missions, e.g. MER \cite{crisp2003mars}, MSL \cite{grotzinger2012mars}, HiRISE \cite{mcewen2007mars}, etc. These datasets have been used to train classifiers to detect classes of interest \cite{panambur2022self,wagstaff2018deep}. Development of annotated datasets allowed for deployment of multi-label Convolutional Neural Networks (CNNs) for classification of both science and engineering tasks for individual missions \cite{lu2021content,wagstaff2021mars} which showed performance improvements over Support Vector Machine (SVM) classifiers. Additional planetary applications include the Science Captioning of Terrain Images (SCOTI) \cite{qiu2020scoti} which generates text captions for terrain images.
%and even rover landing estimation \cite{cheng2005mer} for orientation and acceleration corrections.

\begin{comment}
\begin{itemize}
    \item HIRISENet and MSLNet and MERNet
    \begin{enumerate}
        \item Mars image content classification: Three years of nasa deployment and recent advances \cite{wagstaff2021mars}
        \item Content-Based Classification of Mars Exploration Rover Pancam Images \cite{lu2021content}
    \end{enumerate}
    \item Automated detection of geological landforms on Mars using Convolutional Neural Networks \cite{palafox2017automated}
    \item Deep Mars: CNN Classification of Mars Imagery for the PDS Imaging Atlas \cite{wagstaff2018deep}
    \item SCOTI: Science Captioning of Terrain Images for data prioritization and local image search \cite{qiu2020scoti}
\end{itemize}
\end{comment}

\subsubsection{Terrain Segmentation}
Semantic segmentation for planetary rover missions on Mars has primarily focused on engineering-based tasks such as terrain classification for downstream navigation \cite{cunningham2017locally,higa2019vision,ono2016data}. The development of the annotated AI4Mars dataset \cite{swan2021ai4mars} has enabled the segmentation of Martian surfaces. SPOC (Soil Property and Object Classification) \cite{rothrock2016spoc} leverages the AI4Mars dataset to segment the Martian terrain by utilizing a fully-convolutional neural network. 

\subsubsection{Self-supervised Learning}
Self-supervised and semi-supervised learning can potentially alleviate the significant effort used to annotate rover images in the AI4Mars dataset. Self-supervised networks have aided in the process of annotating images by developing clusters of relevant terrain classes utilizing unlabeled images \cite{panambur2022self}. Contrastive learning has emerged as an important self-supervised learning technique where a network is trained on \textit{unlabeled} data to maximize agreement between randomly augmented views of the same image and minimize agreement between those of different images \cite{chen2020simple,chen2020big}. By using these contrastive-pretrained weights (as opposed to supervised pretrained weights) as a starting point for supervised \textit{finetuning}, it has been shown that contrastive pretraining improves performance on Mars terrain segmentation when only limited annotated images are available \cite{goh2022mars}. This work extends prior work by finetuning the generalized representations obtained through contrastive pretraining on mixed-domain datasets to improve performance across multiple missions.

\section{Methodology}
\label{sec:methods}

\subsection{Dataset Composition}

\begin{figure}[htb]
\centering
\vspace*{-3mm}
\includegraphics[width=1.0\textwidth]{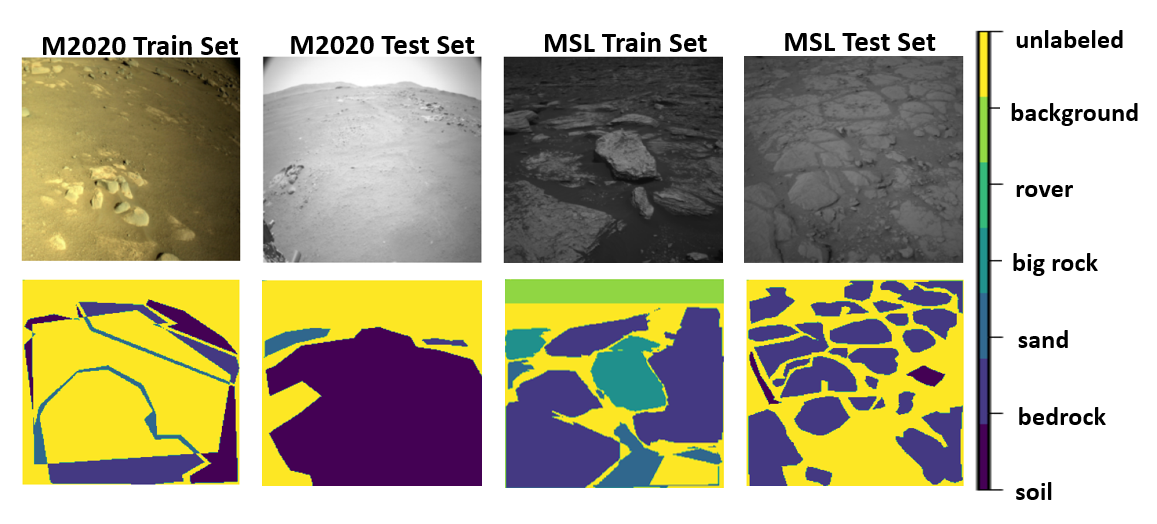}
\caption{Sample images and segmentation masks for M2020 and MSL from the AI4Mars dataset \cite{swan2021ai4mars}. Top row shows the rover captured images and the bottom row shows the corresponding annotations by citizen scientists (train set) or experts (test set).}
\label{fig:sample_imgs}
\vspace*{-3mm}
\end{figure}

\label{sec:dataset}
This section presents the two primary datasets used in this work, as well as the mixed-domain dataset resulting from their combination.
Note that the expert-labeled test sets for MSL and M2020 (322 and 49 images, respectively) are held constant throughout this paper for evaluation. We only manipulate and vary the training set size (number of images) and composition.
\subsubsection{MSL AI4Mars Dataset} 
The AI4Mars dataset \cite{swan2021ai4mars} is comprised of more than 50k images from the PDS captured by the MSL's grey-scale navigation camera (NAVCAM) and color mast camera (Mastcam) as well as MER. 
Citizen scientists around the world annotated MSL images at the pixel-level with one of four class labels, i.e., soil, bedrock, sand, and big rock, resulting in more than 16,000 annotated training images. A testing set of 322 images was composed by three experts with unanimous pixel-level agreement. 

Additionally, the MSL dataset provides two additional masks --- one for rover portions present in the image and another for pixels beyond 30m from the rover. While \cite{swan2021ai4mars} treats pixels that fall under these two masks as null or unlabeled, this paper investigates the feasibility and utility of treating them as additional classes, namely ``rover'' and ``background'', respectively. Unlabeled pixels, or pixels with insufficient consensus in the citizen scientist annotations, are removed from the metrics calculations in training and evaluation.

\subsubsection{M2020 AI4Mars Dataset}
To evaluate multi-mission segmentation on the M2020 mission, this paper leverages images from M2020's color NAVCAM that were annotated using the same citizen scientist workflow for the MSL AI4Mars dataset. The M2020 dataset consists of a training set of 1,321 color images that span Sols 0-157 with four classes, i.e., soil, bedrock, sand, big rock. The testing set was comprised of 49 grey-scale images from Sols 200-203 manually annotated by an expert.
The M2020 and MSL datasets differ primarily in the lack of rover and range masks for M2020 in addition to the image dimensions (1280 x 960 on M2020 compared to 1024 x 1024 on MSL). 
In addition, M2020 training images are full color whereas MSL is grey-scale. The test sets on both MSL and M2020 are grey-scale.

\subsubsection{Deployment for MSL and M2020}
\begin{figure}[htbp]
    \centering
    \vspace*{-6mm}
     \begin{subfigure}[b]{0.46\textwidth}
         \centering
         \includegraphics[width=\textwidth]{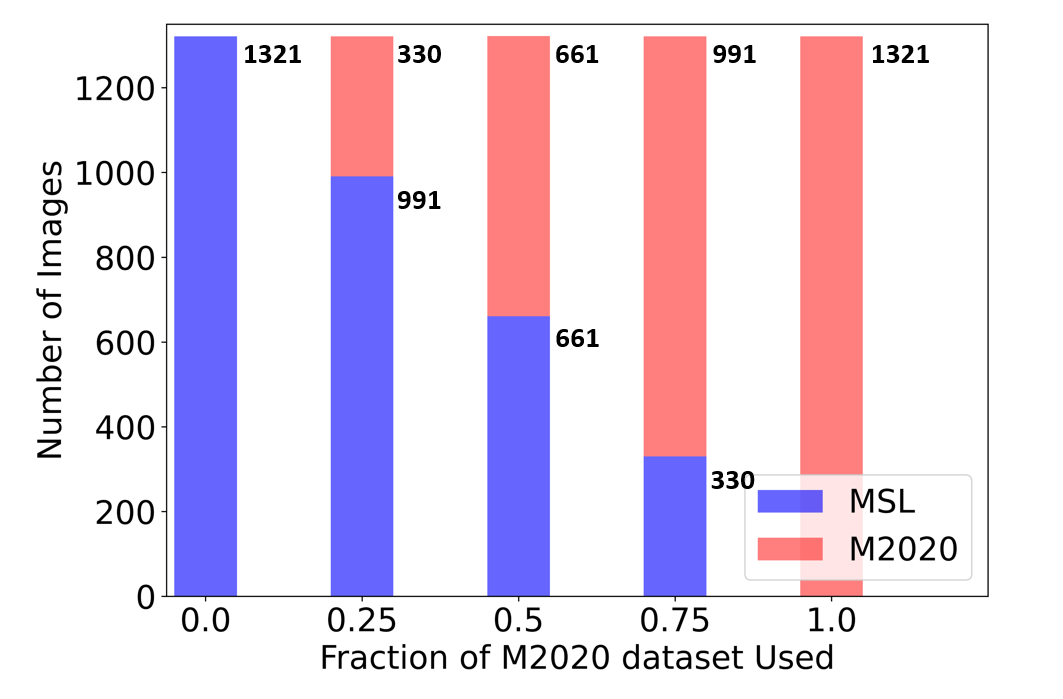}
         \caption{1.3k Images}
         \label{fig:dataset_dist_1k}
     \end{subfigure}
     \hfill
     \begin{subfigure}[b]{0.485\textwidth}
         \centering
         \includegraphics[width=\textwidth]{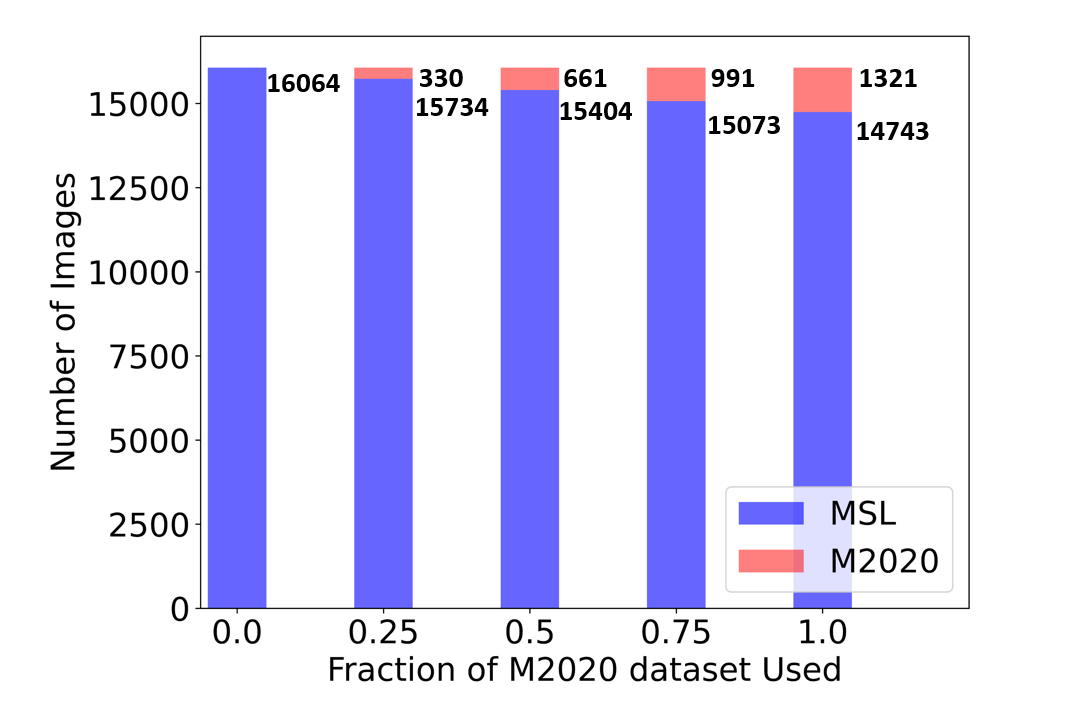}
         \caption{16k Images}
         \label{fig:dataset_dist_16k}
     \end{subfigure}
     \caption{The compositions of the MSL + M2020 mixed datasets across the two training set size experiments, based on different proportions of the M2020 dataset that are present (0.0, 0.25, 0.5, 0.75, 1.0). The number of present images is shown to the right of each bar.}
     \label{fig:dataset_dist}
     \vspace*{-9mm}
\end{figure}

To deploy a single trained model across different missions, we merge the two datasets together, creating a mixed-domain dataset. In doing so, we ensure feature diversity across the domains such that the model does not overfit to one mission. 
To investigate the dependence of our mixed-domain approach on the composition the combined dataset, we performed two different experiments with a different maximum number of images set to match the total number of images in the M2020 training set, 1,321 images (Fig. \ref{fig:dataset_dist_1k}), and the total number of images in the MSL training set, 16,064 images (Fig. \ref{fig:dataset_dist_16k}). 
This maximum number of images allows for the number of images from both the MSL AI4Mars and M2020 datasets to be varied to see the importance of one dataset over the other, since the M2020 dataset is roughly 1/10th the size of the MSL dataset.
% Note the significant difference in number of annotated MSL vs. M2020 images; replacing 1,321 MSL images with all annotated M2020 images still retains 91.8\% of the original MSL Dataset.

\subsection{Semi-Supervised Finetuning}
\label{sec:finetune}
In this work, we train a segmentation network based on the DeepLab architecture \cite{chen2017deeplab} to perform semantic segmentation on Mars terrain images. We finetune pretrained backbones on a combined MSL and M2020 dataset for 50 iterations with a batch size of 16 and a learning rate of $10^{-5}$. To investigate model performance under limited labels, we also vary the number of labeled training images provided from 1\%-100\% of the available 17,385 training images.

% \subsubsection{Architecture}
% The ResNet-101 \cite{he2016deep} model is based on the concept of residual blocks that aid in the training of deep networks. ResNet-101 utilizes 101 layers that can extract both high and low-level features from the image inputs. Ultra-deep networks, previously, became saturated at a certain point due to the problem of vanishing gradients, however, the ResNet architecture overcomes this through shortcut connections \cite{he2016deep}. The ResNet-101 is used as a backbone layer, with pretrained weights transferred and finetuned. Rather than using the fully connected (FC) linear classifier of the ResNet-101 model, a DeepLab Head \cite{chen2017deeplab,chen2017rethinking} is used as a decoder. A DeepLab Head uses Atrous Spatial Pyramid Pooling (ASPP) which captures multi-scale information from feature maps. The use of the DeepLab Head allows a ResNet-101 model that was originally used for image classification to perform semantic segmentation.

% \subsubsection{Contrastive Pretraining}    
The DeepLab architecture combines a ResNet encoder \cite{he2016deep} with a segmentation module based on Atrous Spatial Pyramid Pooling (ASPP) \cite{chen2017deeplab}. Due to the relatively small number of labeled images, on the order of 10k, we leverage transfer learning rather than training the segmentation model from scratch. In other words, we initialize the encoder with pretrained weights and randomly initialize the decoder weights before finetuning for 50 additional epochs on the labeled Mars images. 

The choice of pretrained weights used to initialize the encoder has been shown to play an important role in downstream segmentation performance. In particular, transferring ResNet weights that were pretrained using the SimCLR framework on unlabeled ImageNet images \cite{chen2020big} enables better Mars terrain segmentation performance under limited labels compared to ResNet weights that were pretrained in a supervised fashion (i.e., with labels) on ImageNet \cite{goh2022mars}. In this work, we investigate whether the contrastive pretraining framework is advantageous and generalizes weights for different missions (M2020 vs. MSL) on a mixed domain dataset.

\section{Results}
\label{sec:result}
\subsection{Analysis of Dataset Composition}

\subsubsection{Mixed-Domain Training Improves Multi-Mission Segmentation Accuracy}
% Start with (SPOC) baseline: supervised transfer learning on ResNet-101 on MSL and M2020 separately (i.e., 16,064 vs. 1,321 labels); no need to show % labels variation, just show 100% labels results (accuracy + F1 + confusion matrix). Talk about performance degradation and potential reasons (e.g., train & test sets too small)

Table \ref{tab:baselines} compares segmentation results between supervised and contrastive pretraining methods for the MSL and M2020 evaluation sets. The supervised pretraining tests are referred to as SPOC because they follow \cite{atha2022multi,rothrock2016spoc,swan2021ai4mars} in initializing the encoder using weights that were pretrained in a supervised manner to perform classification on the ImageNet dataset. 
The contrastive pretraining results are referred to as SimCLR as they leverage the weights \cite{chen2020simple,chen2020big} that were pretrained in an \textit{unsupervised} manner on unlabeled ImageNet images. The MSL, M2020, and MSL combined with M2020 were the training sets used to evaluate performance across MSL and M2020 test sets for both pretraining methods. 
\begin{comment}
\begin{table}[htbp]
\vspace*{-6mm}
\centering
\caption{Baseline supervised transfer learning (SPOC) compared with semi-supervised fine-tuning results (SimCLR) across MSL and M2020 segmentation tasks. Combining MSL and M2020 training images provides significant performance improvement on M2020 and a modest improvement on MSL. The SimCLR fine-tuning framework provides a 5\% accuracy improvement on M2020 and a 0.5\% improvement on MSL across both same-mission and combined dataset training.}
\label{tab:baselines}
\begin{tabular}{@{}rllllllll@{}}
\toprule
Trained on: & \multicolumn{2}{c}{\begin{tabular}[c]{@{}c@{}}MSL \\ (16,064 images)\end{tabular}} & \multicolumn{2}{c}{\begin{tabular}[c]{@{}c@{}}M2020 \\ (1,321 images)\end{tabular}} & \multicolumn{4}{c}{\begin{tabular}[c]{@{}c@{}}MSL + M2020\\ (17,385 images)\end{tabular}} \\ \midrule
Evaluated on: & \multicolumn{2}{c}{MSL} & \multicolumn{2}{c}{M2020} & \multicolumn{2}{c}{MSL} & \multicolumn{2}{c}{M2020} \\ \midrule
 & Acc & F1 & Acc & F1 & Acc & F1 & Acc & F1 \\
\multicolumn{1}{l}{SPOC} & 0.959 & 0.840 & 0.662 & 0.495 & 0.965 & 0.860 & 0.745 & 0.544 \\
\multicolumn{1}{l}{SimCLR} & 0.965 & 0.900 & 0.732 & 0.539 & \textbf{0.970} & 0.885 & \textbf{0.796} & 0.610 \\ \bottomrule
\end{tabular}
\vspace*{-3mm}
\end{table}
\end{comment}
%\begin{comment}
\begin{table}[htbp]
\vspace*{-6mm}
\centering
\caption{Baseline supervised transfer learning (SPOC) compared with semi-supervised fine-tuning results (SimCLR) across MSL and M2020. Combining MSL and M2020 training images provides significant performance improvement on M2020 and a modest improvement on MSL.}
\label{tab:baselines}
\begin{tabular}{lcccccccc}
\toprule
Evaluated On: & \multicolumn{4}{c}{MSL} & \multicolumn{4}{c}{M2020} \\ \midrule
Trained On: & \multicolumn{2}{c}{\begin{tabular}[c]{@{}c@{}}In-Domain\\ (16,064 Images)\end{tabular}} & \multicolumn{2}{c}{\begin{tabular}[c]{@{}c@{}}Mixed Domain\\ (17,385 Images)\end{tabular}} & \multicolumn{2}{c}{\begin{tabular}[c]{@{}c@{}}In-Domain\\ (1,321 Images)\end{tabular}} & \multicolumn{2}{c}{\begin{tabular}[c]{@{}c@{}}Mixed Domain\\ (17,385 Images)\end{tabular}} \\ \midrule
 & Acc & F1 & Acc & F1 & Acc & F1 & Acc & F1 \\
SPOC & 0.959 & 0.840 & 0.965 & 0.860 & 0.662 & 0.495 & 0.745 & 0.544 \\
SimCLR & 0.965 & 0.900 & \textbf{0.970} & 0.885 & 0.732 & 0.539 & \textbf{0.796} & 0.610 \\ \bottomrule
\end{tabular}
\vspace*{-3mm}
\end{table}
%\end{comment}
%The SimCLR fine-tuning framework provides a 5\% accuracy improvement on M2020 and a 0.5\% improvement on MSL across both same-mission and combined dataset training.

% Observations from table
% 1. In general, SimCLR > SPOC
%  - MSL: SimCLR acc ~0.5% higher than SPOC
%  - MSL: SimCLR F1 2-6% higher than SPOC
%  - M2020: SimCLR acc 5-7% higher than SPOC
%  - M2020: SimLR F1 4 (M2020 train) - 6% (MSL+M2020) higher than SPOC
% 2. In general, MSL+M2020 outperforms MSL-only/M2020-only finetuning
%  - MSL: MSL+M2020 acc 0.5% higher than MSL-only
%  - MSL: MSL+M2020 F1 -1 (MSL+M2020 actually hurts performance here) to 2% from MSL-only
%  - M2020: 

% \paragraph{Single Domain Training}
When finetuned and evaluated on MSL, a model using either SPOC or SimCLR achieves an average pixel accuracy of ~96\%. However, we see a large performance decrease when the model is finetuned and evaluated on M2020, with an average pixel accuracy 66.2\% and 73.2\% for SPOC and SimCLR, respectively.
This decrease in performance could be attributed to the large discrepancy in dataset sizes. As mentioned in Sec. \ref{sec:dataset}, the number of images in the M2020 dataset is only 8.2\% of MSL (i.e., 1,321 vs. 16,064 images). Furthermore, M2020 only has 49 expert labeled images in the test set, compared to 322 test images in the MSL dataset. 

% \paragraph{Mixed Domain Training}
However, finetuning with a combined dataset of MSL and M2020 images (MSL + M2020 in Table \ref{tab:baselines}) improves performance on both MSL and M2020 test sets across both pretraining methods.
With SimCLR pretraining, finetuning on the combined dataset provides a 6\% accuracy and 7\% F-1 improvement on M2020 compared to finetuning only on M2020 data. On the other hand, SPOC experiences a higher accuracy improvement (8.3\%) and a smaller F-1 improvement (4.9\%) on the M2020 test set when using the combined dataset. 
The improvement on MSL with a combined dataset is marginal, with a 0.5\% increase in accuracy across SPOC and SimCLR when comparing combined dataset finetuning to MSL-only finetuning. For M2020, this performance improvement is driven mainly by the increased number of training examples, which prevents the model from overfitting to only a few M2020 images.

% \paragraph{Impact of Pretraining}
Despite the limited M2020 examples, SimCLR demonstrates higher sample efficiency compared to SPOC, enabling a higher overall pixel accuracy.
Yet, SPOC's larger accuracy improvement could indicate a tendency for the \textit{task-specific} supervised pretrained weights to overfit to classes that are more frequent or more closely related to the ImageNet pretraining task. In contrast, SimCLR's larger F-1 improvement at the cost of accuracy could indicate that the \textit{task-agnostic} contrastive pretrained weights tend to be updated in a more uniform fashion across all classes. In other words, contrastive pretraining acts as a form of implicit regularization in limited-labels regimes such as that encountered by M2020 and enables the model to favor variance in the bias-variance trade-off.

\subsubsection{Model Performance is Sensitive to Images used in Dataset Composition}

\begin{figure}[htb]
     \vspace*{-5mm}
     \centering
     \begin{subfigure}[b]{0.48\textwidth}
         \centering
         \includegraphics[width=\textwidth]{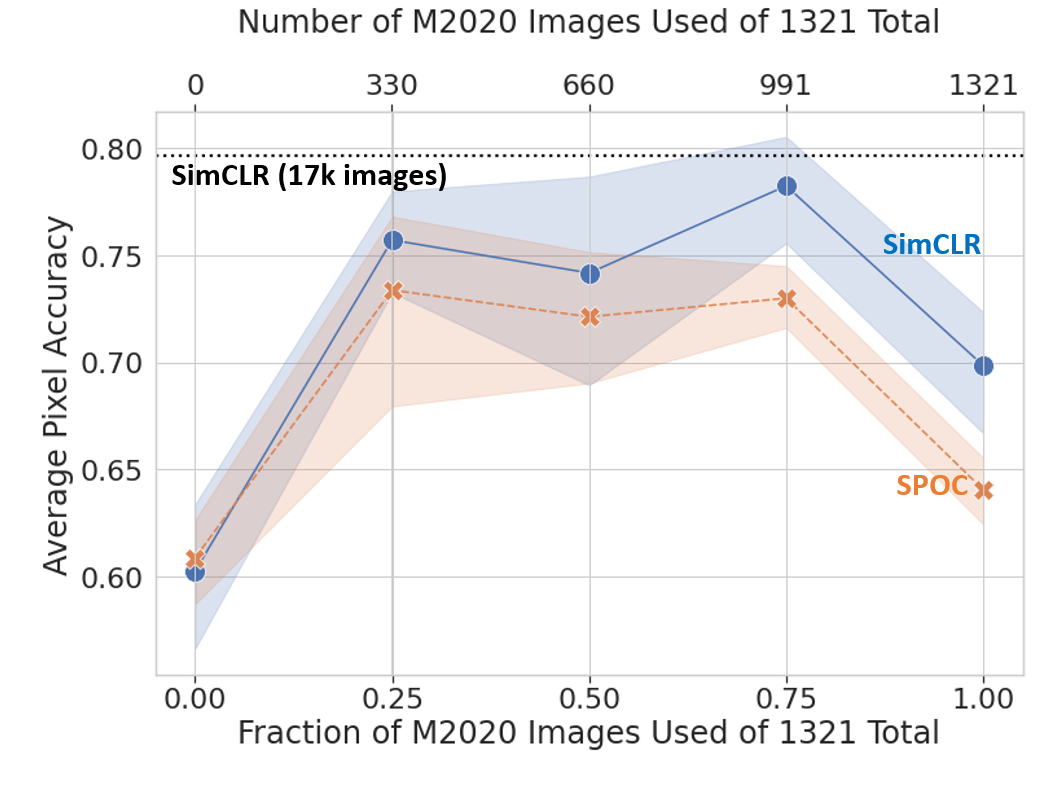}
         \caption{Evaluation on M2020 using 1.3k Imgs}
         \label{fig:m2020_1k_seed}
     \end{subfigure}
     \hfill
     \begin{subfigure}[b]{0.49\textwidth}
         \centering
         \includegraphics[width=\textwidth]{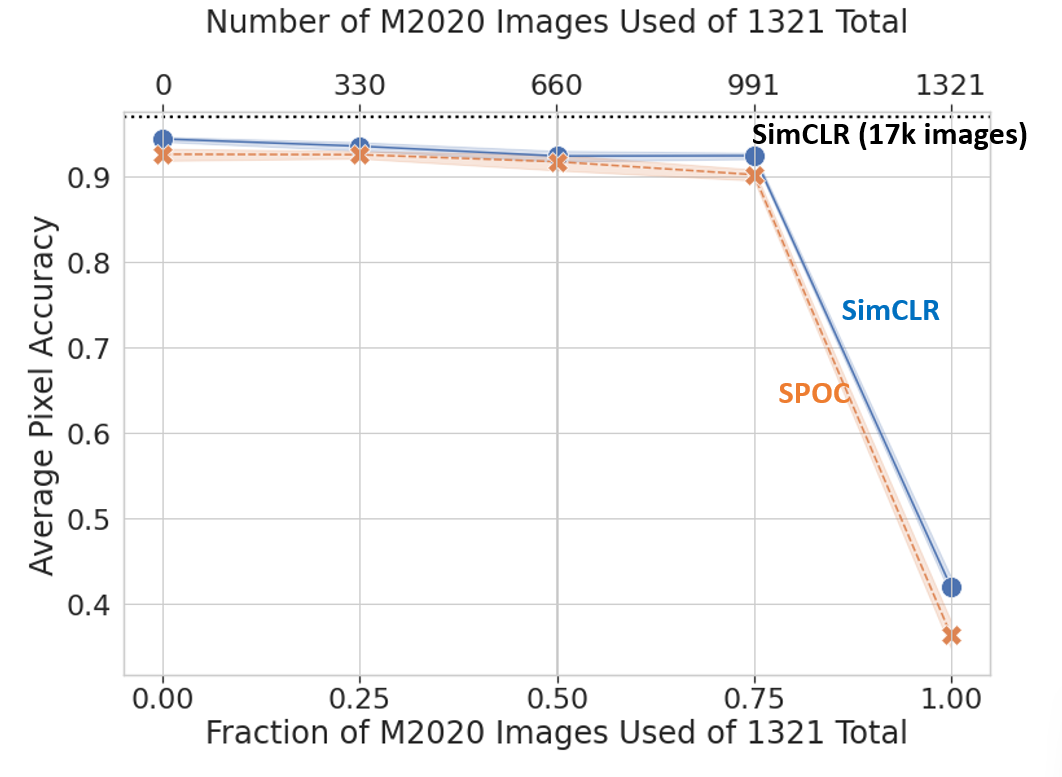}
         \caption{Evaluation on MSL using 1.3k Imgs}
         \label{fig:msl_1k_seed}
     \end{subfigure}
     \caption{Sensitivity of the mixed-domain approach to the combination of MSL images with a specified proportion of M2020 images when the dataset size is capped at 1,321. Solid lines represent the average performance across the different seeds while the bands indicate a 95\% confidence interval.}
     \label{fig:diff_1k_seed}
     \vspace*{-3mm}
\end{figure}

% Varying the % of M2020 used in the combined dataset while keeping 16,064. Combined dataset performs better; why? 
% Run several seeds and report average +/- standard deviation
% seed values for 16k: 10, 42, 63, 95
% seed values for 1.3k: 4, 10, 22, 39, 42, 63, 82, 95, 101, 137
% No COCO, Cross Entropy, and both SPOC and SimCLR
% No variation in num of labels

The initial results (Table \ref{tab:baselines}) establish that using a combined, mixed-domain, dataset for finetuning improves performance on M2020 and MSL across SPOC (ImageNet supervised) and SimCLR (ImageNet contrastive) pretraining. 
This improvement can be attributed to the increased number of training examples (using the combined 17k training set vs. the 16k MSL or the 1.3k M2020 training sets) and the increased \textit{variety} or variance in the domain of training images.
To decouple these potentially-synergistic factors and isolate the mixed-domain effect,  we performed two experiments where we fixed the total number of examples at 1,321 (i.e., the original number of M2020-only labeled images) and 16,064 (i.e., the original number of MSL-only labeled images), then experimented with the number of M2020 images used to replace MSL images in our mixed-domain dataset as illustrated in Fig. \ref{fig:dataset_dist}.
For each dataset split, we ran multiple tests across different seeds to investigate how specific combinations of M2020 images present in the dataset (and conversely, specific combinations of MSL images \textit{not} present) affect model performance. 

Figure \ref{fig:diff_1k_seed} shows the average pixel accuracy across the different sensitivity tests for each fraction of M2020 images used in a total dataset size of 1,321 images. Unsurprisingly, finetuning the SimCLR-pretrained weights with the full 17,385-image MSL+M2020 dataset (79.6\%; black dotted line in Fig. \ref{fig:m2020_1k_seed}) outperforms the average runs across all fractions of M2020 images present.
% from the 16,064-image dataset where 1,321 MSL images are replaced with 100\% of the available M2020 images, albeit by only 3.1\%.
% However, the best run out of the four different seeds used for the 100\% M2020 case actually outperforms the 17,385-image baseline by ~1\%.
% the fortuitous replacement of a specific combination of 1,321 MSL and M2020 images enabled a slight improvement over the best M2020 result in Table \ref{tab:baselines} (81\% vs. 79.6\%) that was obtained using the full 17,385 image dataset. 
However, the performance is improved when there is more \textit{variety} in the training set when compared to an in-domain or complete domain shift training set. This variety from the MSL images could be attributed to the increased number of sites, 77, that MSL explored compared to the 5 sites M2020 explored, thus contributing to performance improvements over the in-domain training.

% Move Fig. 3(c) and 3(d) here, to Fig. 4(a) and 4(b)
\begin{figure}[htbp]
    \vspace*{-3mm}
    \centering
         \begin{subfigure}[b]{0.48\textwidth}
         \centering
         \includegraphics[width=\textwidth]{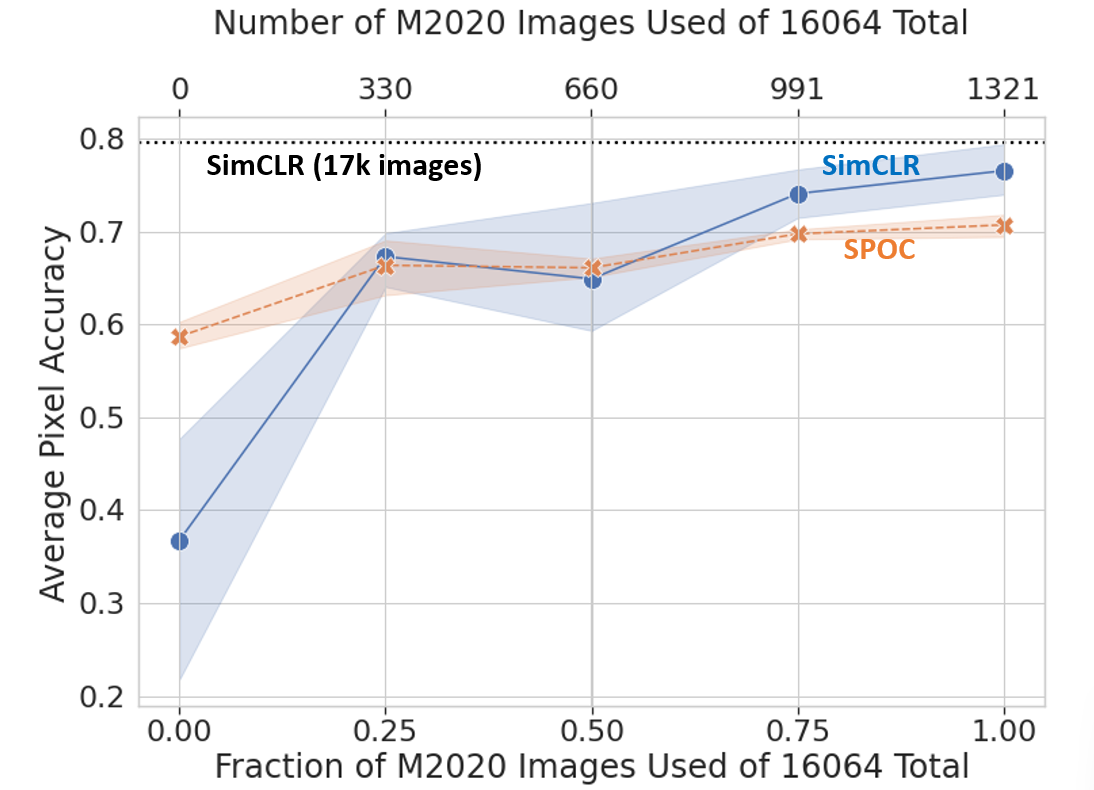}
         \caption{Evaluation on M2020 using 16k Imgs}
         \label{fig:m2020_seed}
     \end{subfigure}
     \hfill
     \begin{subfigure}[b]{0.49\textwidth}
         \centering
         \includegraphics[width=\textwidth]{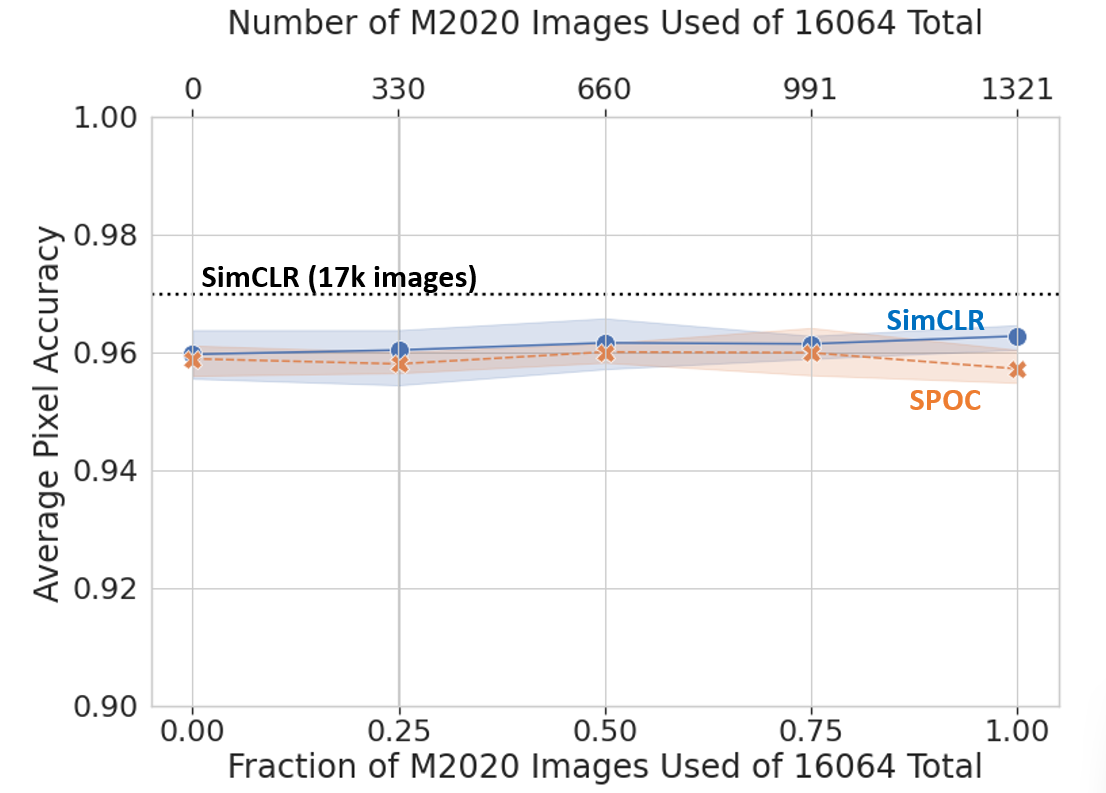}
         \caption{Evaluation on MSL using 16k Imgs}
         \label{fig:msl_seed}
     \end{subfigure}
    \caption{Sensitivity of the mixed-domain approach to the combination of MSL images with a specified proportion of M2020 images when the dataset size is capped at 16,064. Solid lines represent the average performance across the different seeds while the bands indicate a 95\% confidence interval.}
    \label{fig:diff_16k_seed}
    \vspace*{-3mm}
\end{figure}

We see the same improvement when the number of training images is increased to 16k (Fig. \ref{fig:m2020_seed}) since the increased fraction of M2020 images corresponds to a more diverse training set.
This indicates that, independent of the total number of images, the fortuitous replacement of a specific combination of MSL images with M2020 images increases the dataset diversity and is important to the mixed-domain multi-mission terrain segmentation task. 
A larger dataset simply increases the probability that these important images will be seen by the model during training.
% removing certain images from MSL and/or including images from M2020 

% graph a - the MSL dataset is captured from many sols and many sites whereas the M2020 train set is ~100 sols 5 sites and test dataset is from 2 sols and one site - decreased variation in the M2020 dataset
% graph b - checkpoint 1.00 is a complete domain shift, training on M2020 and evaluating on MSL
% graph c - the increased number of M2020 images present in training (replacing MSL images) improves model performance on M2020 test set
% graph d - is the minimal behavior shift because the bulk of the dataset is MSL regardless of how much M2020 is introduced

The sensitivity analysis of different pretraining methods and image samples had a marginal affect on the performance of the MSL test set, so long as there was MSL data present in training. 
Figure \ref{fig:msl_1k_seed} shows that when a model is trained solely on M2020 images then evaluated on MSL, overall performance is decreased by more than 50\% from the 17,385-image baseline.
This significant decrease in performance results from the lack of MSL data present in training as well as the inherent decreased variety from the M2020 dataset that only explored 5 sites.
%However, when MSL images are present in training, there is minimal behavior shift which is expected as the bulk of the mixed-domain datasets, regardless of how many M2020 images are present, is the MSL dataset.
Therefore when MSL images are present in training, there is minimal behavior shift is expected as the MSL image provide more variety in images and make up the bulk of the training set when the data is capped at 16,064 images.
Still, Figures \ref{fig:msl_1k_seed} and \ref{fig:msl_seed} shows across the sampling methods SimCLR outperforms SPOC's supervised pretraining.
% The trend in \ref{fig:m2020_seed} is more sensitive to the proportion of M2020 examples present in training, the increased domain variety, whereas evaluating MSL across different

%\begin{comment}
% Decoder initialization
% \subsubsection{Effects of Different Decoder Initialization}
% The sensitivity analysis experiments were extended to evaluate model performance when introducing a COCO initialized decoder instead of a randomly initialized decoder (Sec. \ref{sec:finetune}). The results showed that even with increased variability, SimCLR pretraining of the encoder outperforms using SPOC supervised pretraining for both random and COCO initialization of the decoder. Yet, the average pixel accuracy across different samples of 100\% M2020 when using the COCO decoder outperformed the average pixel accuracy when using random initialization.

% average pixel accuracy across the different seeds is higher for the coco initialization but there are runs that improve performance with the random initialization (specifically when using SimCLR pretraining). This is likely due to SimCLR being optimized  
%\end{comment}

\subsection{Segmentation with Fewer Labels}

\begin{figure}[htbp]
     \vspace*{-6mm}
     \centering
     \begin{subfigure}[b]{0.49\textwidth}
         \centering
         \includegraphics[width=\textwidth]{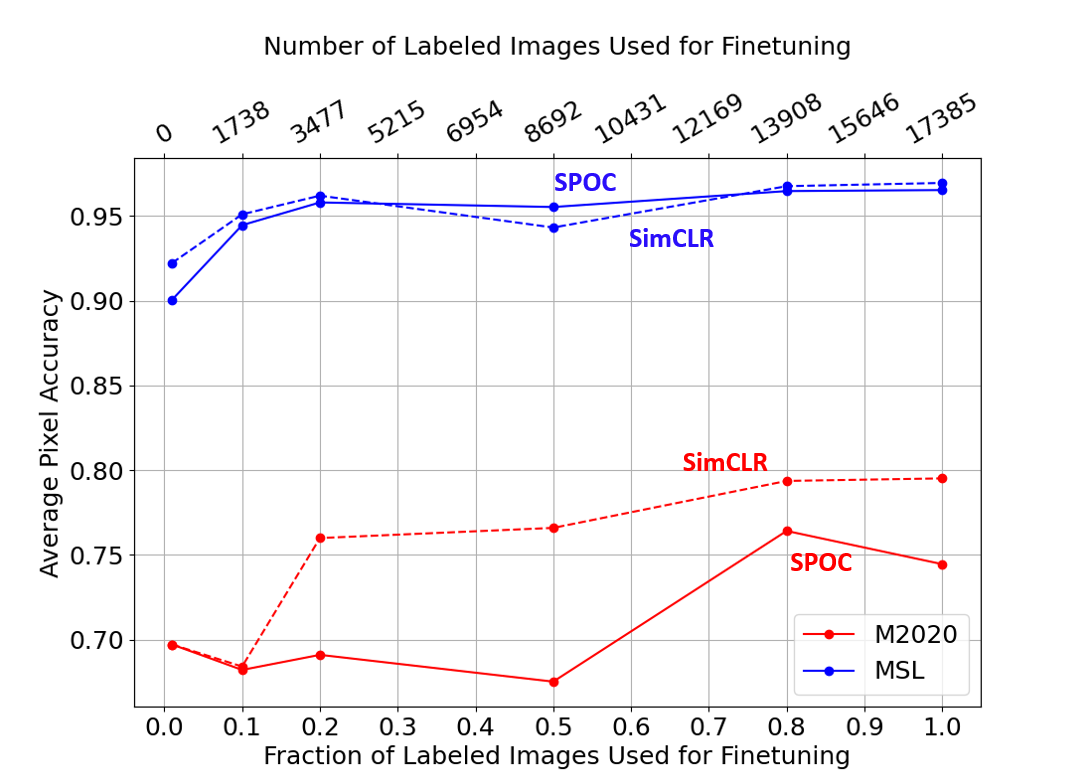}
         \caption{Average Pixel Accuracy}
         \label{fig:label_acc}
     \end{subfigure}
     \hfill
     \begin{subfigure}[b]{0.45\textwidth}
         \centering
         \includegraphics[width=\textwidth]{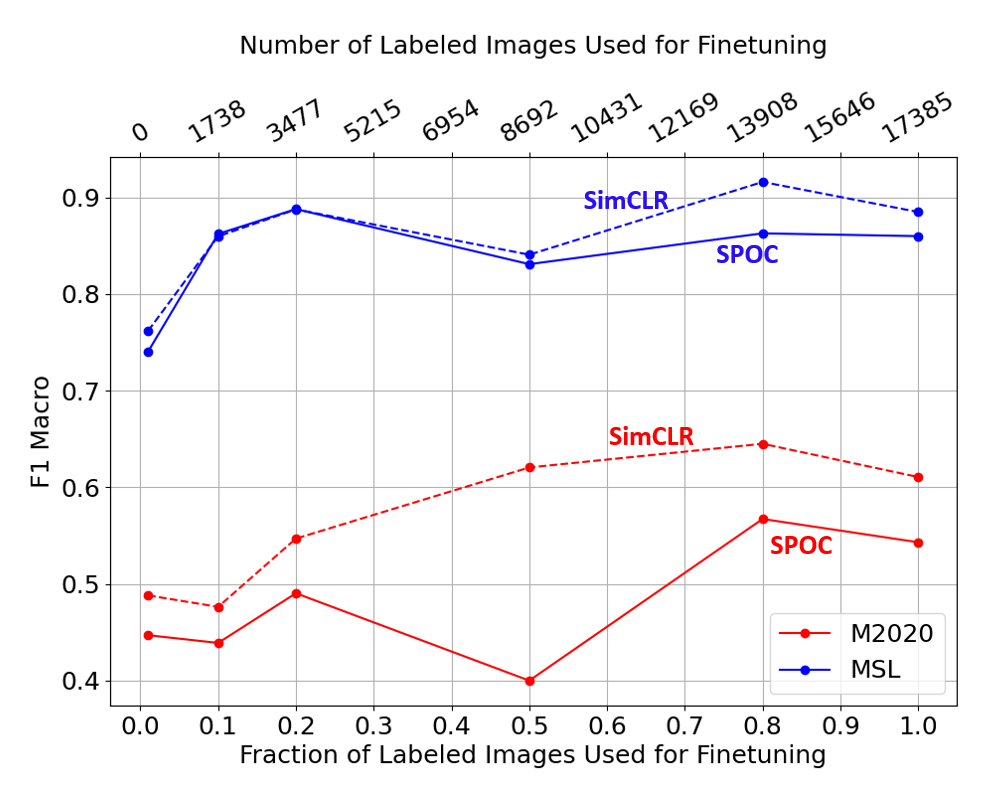}
         \caption{F1 Macro Performance}
         \label{fig:label_f1}
     \end{subfigure}
     \caption{Average Pixel Accuracy and F1-Macro performance vs fraction of labeled images from the mixed-domain dataset used in finetuning the ResNet model. Red lines represent evaluation on the M2020 test set and blue lines represent evaluation on MSL.}
     \label{fig:few_labels}
     \vspace*{-3mm}
\end{figure}
% Varying the number of labels: why we care about varying the number of labels; which combination of model initialization/pretraining + dataset gives the best results
% Use 100% of M2020 + 100% MSL - shown as best run
% Show SPOC vs SimCLR, random decoder initialization

The generation of annotated datasets for the purpose of training deep learning models, specifically from such specific domains as planetary missions, requires significant analysis by experts such as planetary scientists, rover operators, and mission planners. 
This incurs additional time and cost, which can prohibit the application of computer vision-based algorithms on shorter or smaller (i.e., lower-budget) missions.
Thus, utilizing fewer labeled images decreases the need for such efforts. Using the combined training sets from Sec. \ref{sec:dataset}, we vary the number of labeled images used to finetune our network. 
% Varying the distribution of images used in the combined dataset illustrates if adequate performance can be achieved with only a fraction of the available dataset. 
The training datasets will vary in size from 173 (1\% of the available labeled data) to 17,385 (100\% of the available labeled data), note that the expert-labeled test sets are not varied. We stratify the sampled images across the two datasets; the ratio of M2020 to MSL images stays constant across all dataset sizes. For example, a label fraction of 0.1 would contain 1,739 images, of which 8.2\% (143) would be M2020 images while the remaining 91.2\% (1,586) would be MSL images. 

% \subsubsection{Contrastive Pretraining Improves Model Performance with Less Labels}
Figure \ref{fig:few_labels} compares the average pixel accuracy and F1 Macro score achieved by the contrastive (SimCLR) and supervised (SPOC) pretraining methods on both MSL and M2020 test sets when finetuned on varying numbers of labels. 
SimCLR and SPOC perform similarly on the MSL evaluation set, achieving $>$90\% accuracy across the range of label fractions tested. However, SimCLR exhibits a slight accuracy advantage over SPOC below 20\% labels as well as a slight F1 score advantage above 50\% labels.

SimCLR exhibits a larger performance advantage over SPOC on the M2020 test set. At 20\% labels, SimCLR is able to maintain 92.5\% of its full accuracy at 100\% labels (i.e., 3.5\% performance degradation), whereas SPOC experiences a 5.4\% performance decrease.
% Here, the pixel accuracy of M2020 for SimCLR pretraining method continues to improve as more labels are introduced. While highest accuracy for M2020 is when the model is finetuned using 100\% of available labeled images, at 20\% of available labeled images the accuracy is 76\%, which is only a 3.5\% decrease in performance. Additionally, we see that at 50\% of available labeled images, the F1 Macro score (Fig. \ref{fig:label_f1}) has a 1\% increase from using 100\% of labeled images.

\subsection{Ablation Studies}

\subsubsection{Addressing Class Imbalance with Weighted Loss Functions}
\label{sec:loss}
% Loss function - Cross Entropy, Inverse Frequency, Recall

% working on best way to present this information
% since using different loss functions was in an effort to ''rebalance classes'' - metrics like recall and F1 I feel are more important than Average Accuracy 
%   - so long as Accuracy does not decline significantly
% this section will specifically look at M2020 evals 

% TODO: Clean up transition between intro and results

\begin{figure}[htbp]
\vspace*{-6mm}
\centering
\includegraphics[width=0.6\textwidth]{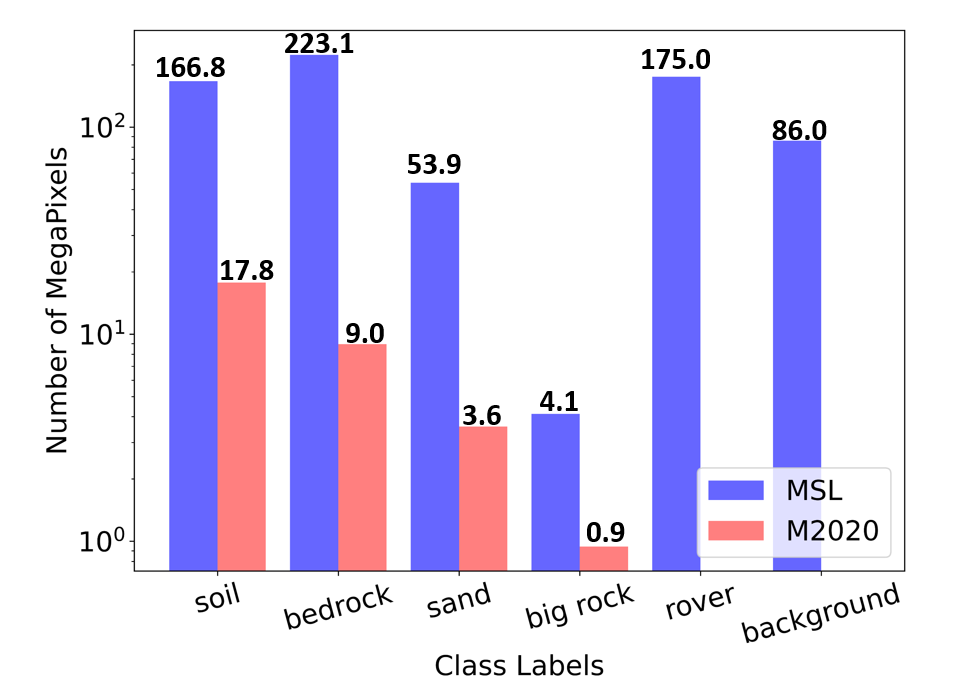}
\caption{Class distribution (in megapixels) of the training sets for M2020 and MSL. Note the absence of rover and background masks in the M2020 dataset.}
\label{fig:class_dist}
\vspace*{-3mm}
\end{figure}

%The number of annotated M2020 images is more accurately only 8.2\% of MSL. 
Class imbalance is an important problem in computer vision tasks such as semantic segmentation because the minority classes are often relevant, e.g., identifying big rocks when navigating rovers. 
Given that the Martian landscape is relatively monotonous and not as label diverse as other standard computer vision datasets \cite{imagenet_cvpr09,lin2014microsoft}, we expect to encounter class imbalance in most planetary datasets.

Figure \ref{fig:class_dist} highlights the class imbalance on the MSL and M2020 datasets. Both datasets exhibit different class distributions --- bedrock is the most frequent terrain element encountered by MSL, whereas soil is the most frequent class encountered by M2020 --- but have the same minority classes of sand and big rock.
% Also the AI4Mars datasets are annotated by multiple citizen scientists or experts and labels only become valid after a specified number of annotators have worked on the given image and their annotations reach a certain agreement threshold. This agreement threshold is used to enhance overall set quality by reviewing the pixel-wise label overlap of the scientists annotations before accepting the label as ground truth. 
% This leads to certain classes like sand and big rock to have less agreement because of their inherent ambiguity \cite{swan2021ai4mars} as compared to soil or bedrock.
% Figure \ref{fig:class_dist} shows the present class label frequency distributions for the different datasets, i.e. MSL, M2020. 
% The frequencies confirm the expectation that the more ambiguous class, big rock, is less present within the training sets of each dataset.
This section presents investigations into different loss functions aimed at improving performance on minority classes such as big rock and sand. 

\paragraph{Cross-entropy Loss}
Most of the existing work in semantic segmentation utilize the cross-entropy loss function. 
This loss function is widely used in classification tasks not only because it encourages the model to output class probability distributions that match the ground truth, but also because of its mathematical properties that enable efficient backpropagation during the training of neural networks. 
Following the derivation in \cite{tian2021striking}, the cross-entropy loss can be written in the form given in Eq. \ref{eq:cross_entropy},

\vspace*{-3mm}
\begin{equation}
    CE = -\sum_{c=1}^C{N_c \log{P_c}}
    \label{eq:cross_entropy}
\end{equation}
where $c \in {1, ..., C}$ is the class among a set of $C$ classes, $N_c$ is the number of examples that belong to class $c$, and $P_c = \left(\prod_{n:y_n=c}^{N_c}{P_n^c}\right)^{1/N_c}$ represents the model's geometric mean confidence across all examples belonging to class $c$.
Equation \ref{eq:cross_entropy} highlights that a ``vanilla'' cross-entropy loss implementation biases towards classes with larger $N_c$ for imbalanced datasets, and could reduce model performance as it does not consider the probability mass distribution across all classes \cite{grandini2020metrics}. 
Thus, we experimented with several loss functions to improve overall model performance as well as performance on individual classes.

\vspace*{-3mm}
\begin{align}
\label{eq:weighted}
InvCE &= -\sum_{c=1}^C\frac{1}{freq(c)}N_c\log(P_c) \\
 &= -\sum_{c=1}^C\frac{1}{N_c}N_c\log(P_c) = -\sum_{c=1}^C\log(P_c)
\end{align}

\paragraph{Inverse Frequency Loss}
A common method of handling class imbalance is to weight the loss function so as to prioritize specific classes. The most common weighting method is to use inverse frequency, which assigns more importance to minority/rare classes (i.e., classes with few examples) \cite{tian2021striking}.
The frequency of a class is found by dividing the number of pixels in a class c, $N_{c}$, by the total number of pixels, N. Inverse frequency weighting can over-weight the minority classes, so to reduce this effect and the addition of false positives, we normalize the weights. 

%add explanations of the variables used in each equation. also, we need to make sure that the subscript / superscript conventions in all equations are consistent.

\paragraph{Recall Loss}

The recall metric measures a model's predictive accuracy --- it is given by $\frac{TP}{TP + FN}$, which is the number of true positives (i.e., pixels that the model \textit{correctly predicted} as belonging to the positive class) divided by the sum of true positives and false negatives (i.e., the total number of pixels that actually belong to the positive class).
%the fraction of true positive $(TP)$ predictions divided by the total number of positively classified pixels (the sum of the true positive and false negative $(FN)$ predictions) \cite{grandini2020metrics}.
Recall loss \cite{tian2021striking} is a novel method that weights a class based on the training recall metric performance as seen in Eq. \ref{eq:recall} and even extends the concept of inverse frequency weighting. The true positive and false negative predictions are used with respect to class $c \in 1, ..., C$ and the time dependent optimization step, $t$, to optimize the geometric mean overall classes at $t$, $p_{n,t}$.

\vspace*{-3mm}
\begin{equation}
\begin{split}
\label{eq:recall}
RecallCE &= -\sum_{c=1}^C(1-\frac{TP_{c,t}}{FN_{c,t}+TP_{c,t}})N_c\log(p^{c,t}) \\
&= -\sum_{c=1}^C\sum_{n:y_i=c}(1-R_{c,t})\log(p_{n,t})
\end{split}
\end{equation}

Recall loss is weighted by the class-wise false negative rate. We expect the model to produce more false negatives (i.e., exhibit low recall) on rare classes that are more challenging to classify, whereas more prevalent classes will produce fewer false negatives. This allows the recall loss to dynamically favor classes with lower per-class recall performance $(R_{c,t})$ during training.
Recall loss has been shown to improve accuracy while maintaining a competitive mean Intersection-over-Union (IoU).

% Section \ref{sec:dataset} discusses the class imbalance across both the MSL and M2020 datasets. Given that minority classes such as big rock and sand are often hazards that inhibit mobility, the inability to correctly predict these classes could increase risk when deployed on a rover. 
% Thus looking at overall pixel accuracy does not reflect the performance across individual classes. 
% For rover navigation it is important to identify the terrains that are traversable (e.g. soil and bedrock) and detect potential dangers (e.g. sand and big rock).  
% Using the different loss weighting methods, Inverse Frequency and Recall Cross Entropy, allows one to find a balance between accuracy, recall, and precision when training a model. 

\begin{comment}
\begin{table}[htbp]
%(Pixel Accuracy, F1 Macro, mIoU, and minority class, Big Rock Recall) 
\vspace*{-6mm}
\centering
\caption{Performance metrics across different different loss functions on M2020 evaluation.}
\label{tab:loss_funcs}
\begin{tabular}{l|cccc}
\toprule
Loss Function & Accuracy & F1 Macro & mIoU & Big Rock Recall \\ \midrule
Cross Entropy & 0.742 & 0.573 & 0.429 & 0.1689 \\
Inverse Frequency & 0.774 & 0.587 & 0.455 & 0.454 \\
Recall CE & 0.759 & 0.599 & 0.153 & 0.502 \\ \midrule
\begin{tabular}[c]{@{}l@{}}\textbf{Inverse Frequency} \\ \textbf{+ Recall CE}\end{tabular} & 0.765 & 0.571 & 0.450 & \textbf{0.550} \\ \bottomrule
\end{tabular}
\vspace*{-3mm}
\end{table}
\end{comment}

\begin{table}[htbp]
\vspace*{-6mm}
\centering
\caption{Performance metrics across different different loss functions on M2020 evaluation.}
\label{tab:loss_funcs}
\begin{tabular}{l|cccc}
\toprule
Loss Function & Accuracy & F1 Macro & mIoU  & Big Rock Recall \\ \midrule
Cross Entropy & 0.795 & 0.610 & 0.480 & 0.120 \\
Inverse Frequency & 0.757 & 0.582 & 0.449 & 0.350 \\
Recall CE & 0.781 & 0.607 & 0.474 & 0.164 \\ \midrule
\begin{tabular}[c]{@{}l@{}} \textbf{Inverse Frequency}\\ \textbf{+ Recall CE} \end{tabular}& 0.772 & 0.591 & 0.458 & \textbf{0.480} \\ \bottomrule
\end{tabular}
\vspace*{-3mm}
\end{table}

When evaluating model performance on an imbalanced dataset, overall pixel accuracy does not provide adequate insight into the model's performance on individual classes. As such, we use the F1 score, which is the harmonic mean of precision and recall. The F1 score takes into account both the false positives and false negatives, whereas recall only accounts for the false positives along with the true positives. Furthermore, we show the mean intersection over union (mIoU), which is the percent of overlap in the prediction and ground truth. 

\begin{figure}[htbp]
    %\vspace*{-3mm}
     \centering
     \begin{subfigure}[b]{0.45\textwidth}
         \centering
         \includegraphics[width=\textwidth]{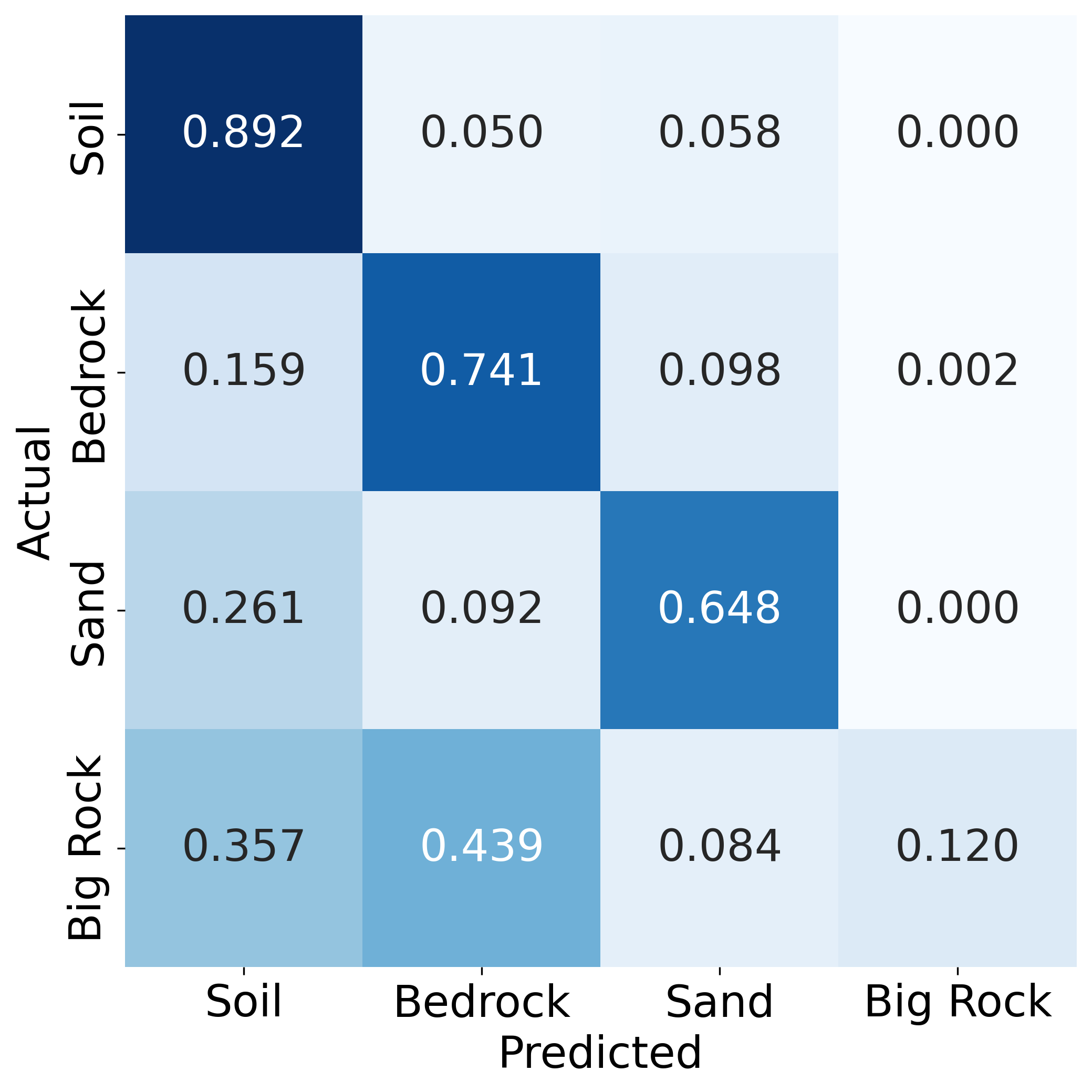}
         \caption{Cross Entropy -- Acc: 79.5\%}
         \label{fig:base_cm}
     \end{subfigure}
     \hfill
     \begin{subfigure}[b]{0.45\textwidth}
         \centering
         \includegraphics[width=\textwidth]{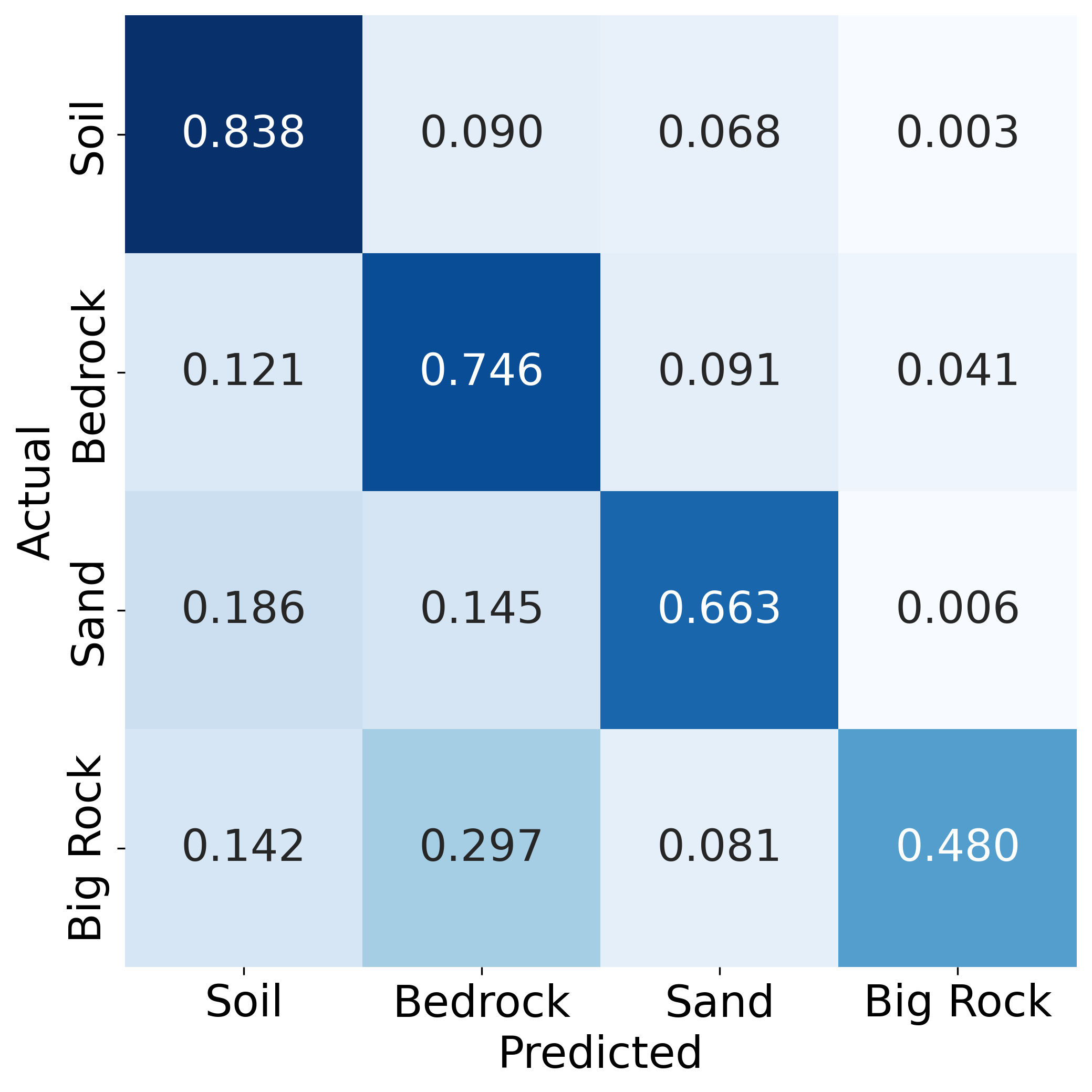}
         \caption{Inv Freq + Recall CE -- Acc: 77.2\%}
         \label{fig:loss_ablation_cm}
     \end{subfigure}
     \caption{Confusion matrices for Cross Entropy loss and Inverse Frequency + Recall Cross Entropy loss as a result of the ablation studies. The confusion matrix shows a 36\% increase in ability to positively predict big rock using Inv Freq + Recall CE loss compared to Cross Entropy loss.}
     \label{fig:confusion_matrices}
     \vspace*{-3mm}
\end{figure}

Table \ref{tab:loss_funcs} compares the performance utilizing different loss functions when finetuning a contrastive-pretrained ResNet-101 model on the 17,385-image mixed-domain training set.%that uses 100\% M2020 and 100\% MSL. 
Results show that weighting the loss function improves the minority class positive predictions, pixel accuracy, and the F1 Macro score. In particular, Recall Cross Entropy (CE) alone improves Big Rock recall but decreases mIoU value when compared to the baseline CE loss. The combined Recall CE and Inverse Frequency loss have the highest recall on the Big Rock class with a value of 0.48. 
% Unlike when using Recall CE alone, overall model performance was not negatively affected and increased by 2\% when compared with CE Loss.
% Further combining the two weighted loss functions continues to improve the Big Rock Recall to a value of 0.55, while also increasing pixel accuracy by 2\% compared to using Cross Entropy loss.
The confusion matrices in Fig. \ref{fig:confusion_matrices} better visualize the ``rebalancing'' of classes when using the combined loss functions over the CE loss. Figure \ref{fig:loss_ablation_cm} shows that the combined weighted loss function was able to improve correct predictions of big rock over the cross-entropy loss. 
Additionally, Cross-entropy loss model tends to over predict soil since it is the most frequent class (Fig. \ref{fig:base_cm}), but Inv. Freq and Recall is able to address all classes more uniformly at a small expense of lower Soil recall. 

%In particular, Fig. \ref{fig:loss_ablation_cm} shows that not only did the model increase the ability to positively predict big rock but also sand; in turn reducing the number of false negative predictions.

\subsubsection{Bigger Models Increase Performance}
% Model size experiments - let's just do it with pure MSL, evaluate on MSL
%   - compare: resnet50, resnet101, mobilenetv2, resnet101_2x_sk1 (257M++ params), resnet101_2x_sk1 intermediate layer connect to SegmentationHead
% Comparing 6 classes vs. 4 classes

% how to present rather large table?
% drop intermediate layer from data
% use to color map across number of parameters
%   - darker is more num of parameters
% is difference better in F1? 

\begin{figure}[htbp]
     \centering
     %\vspace*{-3mm}
     \begin{subfigure}[b]{0.475\textwidth}
         \centering
         \includegraphics[width=\textwidth]{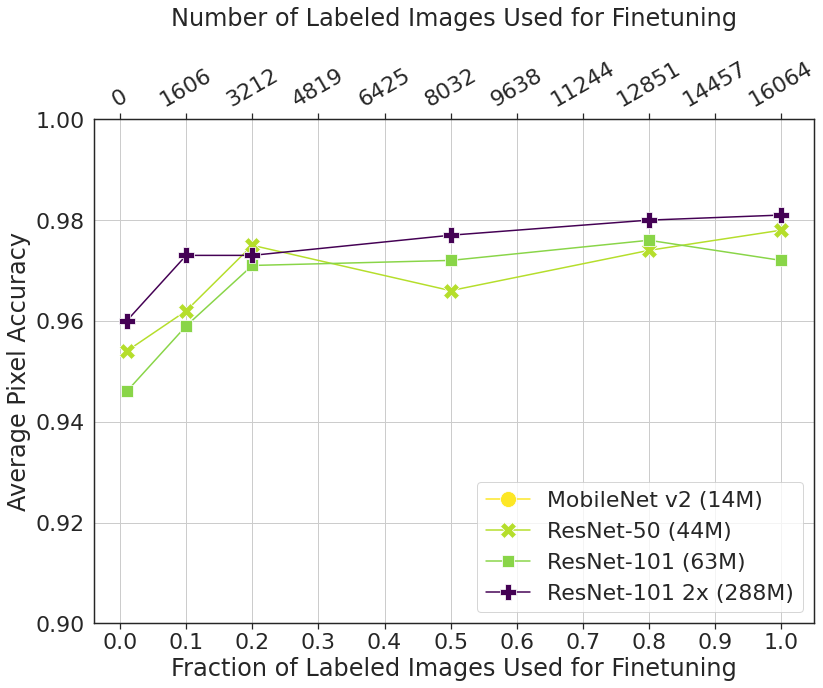}
         \caption{SimCLR Pretraining}
         \label{fig:large_simclr}
     \end{subfigure}
     \hfill
     \begin{subfigure}[b]{0.475\textwidth}
         \centering
         \includegraphics[width=\textwidth]{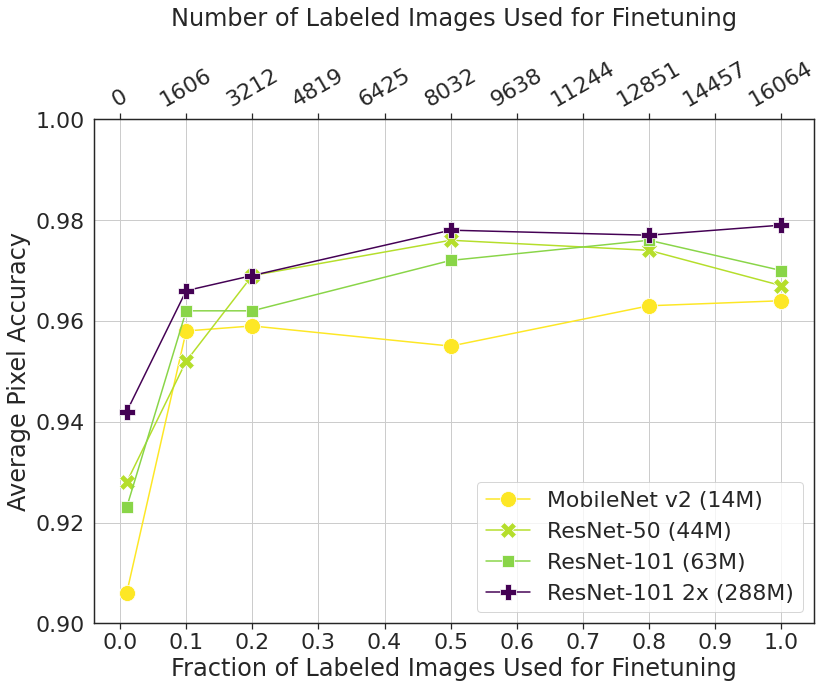}
         \caption{SPOC Pretraining}
         \label{fig:large_spoc}
     \end{subfigure}
     \caption{Comparison of SPOC and SimCLR across different sized models when finetuned on a varying number of labeled images from the MSL dataset and evaluated on the MSL test set. Note that contrastive pretraining weights are not available for the MobileNet v2 model.}
     \label{fig:large_model}
     \vspace*{-3mm}
\end{figure}

To evaluate the importance of model size, specifically the number of parameters, we simplified our experiments. Four different models, MobileNet v2, ResNet-50, ResNet-101, and ResNet-101 with 2x width, were used to perform finetuning with a training set of the MSL dataset (16,064 images). Figure \ref{fig:large_model} shows that across a varied number of labeled images used in finetuning, that larger models with an increased number of parameters have the highest pixel accuracy. In comparison to the MobileNet v2 (14 M parameters), the ResNet-101 2x (288 M parameters) improves accuracy at 1\% of the available labeled images by 5\% and when using 100\% of the available labeled images there is ~2\% increase. 
Additionally, it is shown that when using these larger model sizes (e.g. ResNet-101 and ResNet-101 2x) with fewer annotated images improves accuracy over smaller models (e.g. MobileNet v2, ResNet-50), but the increase in accuracy with the increase of available labels starts to diminish. These diminishing returns in pixel accuracy question the trade-off in compute expense and accuracy specifically on a constrained edge-device like a planetary rover.
%Additionally, when using SimCLR pretraining, the larger models (ResNet-101 and ResNet-101 2x) maintain the highest F1 scores across the varied number of annotated images.

\section{Conclusion}

This research proposes a state-of-the-art semi-supervised mixed-domain training approach for multi-mission Mars terrain segmentation by combining annotated MSL images from the novel AI4Mars dataset with newly-annotated M2020 images to finetune a contrastive pretrained model, thereby improving performance across both MSL and M2020 test sets.

%for a mixed-domain finetuning of a self-supervised model which improved upon results across both MSL and M2020 datasets. 
The idea of combined datasets is not novel, yet decoupling the effects of change in domain training and change in number of samples revealed their relative importance within the context of multi-mission deployment. It was found that while the number of images plays a significant role in enabling the performance gain from a combined dataset, the additional variance in a mixed-domain dataset prevents the model from overfitting to unhelpful examples.
% IMPORTANT CAVEAT THAT WE DIDN'T ADDRESS:
% ROVER IMAGES IN M2020. What if we replaced perfectly good MSL images with all problematic ROVER images from M2020
We then show that a self-supervised contrastive pretraining method outperforms the published baseline \cite{rothrock2016spoc} by 5\%, using only 20\% of the available labels on the combined dataset. 
Finally, adding different class weighting methods aided in addressing the class imbalance issue that is inherently present within the planetary datasets. By improving performance on the minority classes (i.e. big rock and sand), we can improve a rover's ability to detect and avoid these potentially hazardous areas.

Areas of future work include expanding this approach to handle multi-task learning (e.g. scientific vs engineering class sets). The model would become valuable as a mission- and task-agnostic method for semantic segmentation in planetary explorations. Additionally, evaluating the performance of the larger model sizes for a multi-mission effort will aid in the development of a distillation pipeline for the deployment of a smaller, computationally less expensive model on a planetary rover.

% would sample predictions be valuable in the conclusions or in the baselines?
%   - also need to identify best looking predictions
%   - is there enough room?

% Contribution 1. multi-mission - combined datasets isn't novel, but we're adding to the understanding regarding mixed-domain training; isolated change in domain from change in num. labels
% Contribution 2. introduce self-supervised learning SimCLR - improves performance on label-limited M2020 dataset by 5% over published baseline; exceeds published baseline performance with only 20% of the labels
% Contribution 3. addressing class imbalance with different loss functions
 
% Larger models produce better performance on MSL when trained on MSL
% ResNet-101 2x sk1 is better performance can we use it as a teacher for a much smaller model like MobileNetv2 as part of distillation

\section{Acknowledgments}
This research was carried out at the Jet Propulsion Laboratory, California Institute of Technology, under a contract with the National Aeronautics and Space Administration (80NM0018D0004), and was funded by the Data Science Working Group (DSWG). The authors also acknowledge the Extreme Science and Engineering Discovery Environment (XSEDE) Bridges at Pittsburgh Supercomputing Center for providing GPU resources through allocation TG-CIS220027. The authors would like to thank Hiro Ono and Michael Swan for facilitating access to AI4Mars. U.S. Government sponsorship acknowledged.

\bibliographystyle{splncs04}
\bibliography{egbib}
\end{document}